\documentclass[review]{elsarticle}
\PassOptionsToPackage{hyphens}{url}
\usepackage{hyperref}

\usepackage{latexsym} 
\usepackage{graphics}
\usepackage{epsfig}
\usepackage{graphicx}
\usepackage{tikz-qtree}
\usepackage{tikz}
\usepackage{amsmath}
\usepackage{dirtytalk} 
\usepackage{bbm} 
\usepackage{algorithm}
\usepackage{algpseudocode}
\usepackage{pgfplots}
\usepgfplotslibrary{groupplots}
\usepackage[utf8]{inputenc}
\usepackage[english]{babel}
\usepackage{graphicx}
\usepackage{subcaption}
\usepackage{tikz}
\usepackage{graphicx}
\usepackage{subcaption}
\usepackage{pgfplots}
\pgfplotsset{width=11cm,compat=newest}
\usepackage{algorithm}
\usepackage{booktabs}
\makeatletter
\def\BState{\State\hskip-\ALG@thistlm}
\makeatother
\algdef{SE}[DOWHILE]{Do}{doWhile}{\algorithmicdo}[1]{\algorithmicwhile\ #1}%

\usepackage{varwidth}
\usepackage{booktabs} 
\usepackage{epsfig}
\usepackage{amsfonts}
\usepackage{array}

\usepackage{rotating}
\usepackage{tabularx}
\usepackage{color}
\usepackage{multirow}
\usepackage{url}
\usepackage{lscape}
\usepackage{xargs} 
\newcolumntype{x}[1]{>{\centering\let\newline\\\arraybackslash\hspace{0pt}}p{#1}}
\usepackage{pgfplots}
\pgfplotsset{compat=1.17}

\usepackage{tabularx}
\usepackage{array}
\usepackage{colortbl}
\usepackage{multirow}
\usepackage{datetime}
\usepackage[mathscr]{eucal}
\usepackage{amssymb}
\usepackage{amsmath}
\usepackage{graphicx}
\usepackage{derivative}
\usepackage[utf8]{inputenc}
\usepackage[T1]{fontenc}
\usepackage{mathtools}
\usepackage[thinc]{esdiff}
\usepackage{wrapfig}
\usepackage{caption}
\usepackage{lscape}
\usepackage{rotating}
\usepackage{url}
\usepackage{babel}
\usepackage{translator}
\usepackage{pgfkeys,pgfcalendar}
\usepackage{graphicx}
\usepackage{multirow}
\usepackage{hhline}
\usepackage{diagbox}
\usepackage{eurosym}
\usepackage{amssymb}
\usepackage{pifont}
\newcommand{\cmark}{\ding{51}}%
\newcommand{\xmark}{\ding{55}}%
\graphicspath{ {/home/ciaran/Downloads/} }
\usepackage{acronym}
\usepackage{booktabs}
\def\preparecolorrefs#1{%
  \setcounter{refindex}{0}%
  \whiledo{\value{refindex}<#1}{%
    \stepcounter{refindex}%
    \expandafter\def\csname\therefindex color\endcsname{black}%
  }%
}
\usepackage{hyperref}

\journal{ArXiv}

\biboptions{authoryear}

\begin{document}

\begin{frontmatter}

\title{Electricity Price Forecasting in the Irish Balancing Market}

\author{Ciaran O'Connor}
\address{SFI CRT in Artificial Intelligence, University College Cork, Ireland}
\ead{119226305@umail.ucc.ie }

\author{Joseph Collins}
\address{School of Mathematical Sciences, University College Cork, Ireland}
\ead{98718584@umail.ucc.ie }

\author{Steven Prestwich, Andrea Visentin}
\address{Insight Centre for Data Analytics, University College Cork, Ireland}
\ead{s.prestwich@cs.ucc.ie, andrea.visentin@insight-centre.org}

\begin{abstract}
Short-term electricity markets are becoming more relevant due to less-predictable renewable energy sources, attracting considerable attention from the industry. The balancing market is the closest to real-time and the most volatile among them. Its price forecasting literature is limited, inconsistent and outdated, with few deep learning attempts and no public dataset. This work applies to the Irish balancing market a variety of price prediction techniques proven successful in the widely studied day-ahead market. We compare statistical, machine learning, and deep learning models using a framework that investigates the impact of different training sizes. The framework defines hyperparameters and calibration settings; the dataset and models are made public to ensure reproducibility and to be used as benchmarks for future works. An extensive numerical study shows that well-performing models in the day-ahead market do not perform well in the balancing one, highlighting that these markets are fundamentally different constructs. The best model is LEAR, a statistical approach based on LASSO, which outperforms more complex and computationally demanding approaches.

\end{abstract}


\end{frontmatter}




\section{Introduction}\label{sec:introduction}
Accurate price forecasts are challenging given the characteristics of short-term electricity market prices, i.e. high volatility, sharp price spikes, and seasonal demand. The continuing deployment of renewables and battery energy storage systems is likely to lead to increased price volatility \cite{martinez2016impact, statisticsexplained}. 

The \emph{Balancing Market} (BM) is the last stage for trading electric energy, exhibiting far higher volatility compared to both the \emph{Day-Ahead Market} (DAM) and \emph{Intra Day Market} (IDM). It plays an essential role (in particular in regions where storage of large quantities of electric energy is not economically convenient \cite{MAZZI2017259}) as production and consumption levels must match during the operation of electric power systems. The growing importance of accurate forecasts of BM prices to participants is outlined in \cite{ortner2019future}, where forecast errors of variable renewable electricity will drive demand for BM participation.

Historically, the focus on the DAM is intuitive, given that it is a cornerstone of the European electricity market. In addition, the datasets required for forecasting the DAM are widely available. The lack of analysis of the BM is likely the result of a combination of factors including not all jurisdictions having a BM, the rules governing it can differ from region to region and the identification and acquisition of the relevant datasets can be complicated and expensive (with no open access dataset). 

In recent years, given access to additional datasets and increasing GPU speeds, the application of \emph{Deep Learning} (DL) models has become an attractive option. Short-term \emph{Electricity Price Forecasting} (EPF), in particular, has seen an increasing number of publications pertaining to the DAM, with interest moving away from statistical approaches and towards \emph{Machine Learning} (ML) approaches, \cite{weron2014electricity, neupane2017ensemble, luo2019two}. While DL techniques have gained prominence in recent DAM EPF studies, their evaluation in the context of BM forecasting remains limited in the existing literature. Hence, our emphasis is on creating a benchmark using high-performing models from DAM EPF research to assess the performance of DL methods in the considerably more volatile BM setting.

\subsection{Motivation and contributions}
Accurate price forecasting is of particular interest to generators, buyers, and energy traders, including quick-response participants like battery energy storage systems that mitigate real-time supply-demand volatility. Non-physical financial traders, such as Net Imbalance Volume chasers \cite{NIV}, also capitalize on price spreads between ex-ante markets and the BM.
Recent publications advocate best practices, promoting open-access DAM-related datasets and models \cite{lago2018forecasting, lago2021forecasting}. Building on this, our paper extends the approach to the BM, presenting an analogous framework. Our contribution is to:

\begin{itemize}
    \item Define a replicable framework to train and evaluate forecasting models on the BM. This includes clear instructions on the hyperparameter optimisation for each model, including providing the accompanying dataset. The data and code used for this work are available in \href{https://github.com/ciaranoc123/Balance-Market-Forecast}{GitHub} \footnote{https://github.com/ciaranoc123/Balance-Market-Forecast}.
    
    \item Benchmark the performance of prominent statistical, ML, and DL models, originally designed for the DAM EPF domain, when applied to the BM context. 

    \item Compare model performance between BM and DAM price forecasting, offering insights into handling volatile price movements. 

    \item We explore the unique difficulties that DL models encounter within the significantly more volatile BM. Our investigation reveals that DL models struggle with complex price fluctuations in the BM, resulting in elevated prediction errors.

    \item We conduct a comprehensive analysis to examine how changing the training data size impacts forecast accuracy across various predictive models. 
  
\end{itemize}

The rest of this paper is structured as follows.
In Section \ref{epfliteraturereview}, we refer to some of the more recent \& salient DAM and BM EPF publications. Section \ref{electricitymarketstructure} provides an overview of the structure of a typical European electricity market and describes the peculiarities of the Irish BM. In Section \ref{analyticalframework}, we present the specifics of the forecasting problem and the associated analytical framework.
Section \ref{modelssec} describes the models utilised in the paper. The experimental results and their analysis are presented in \ref{resultssec}. Finally, \ref{conclusionsec} concludes the paper with observations and future works.

\section{Literature Review}\label{epfliteraturereview}
We highlight some of the more recent and salient short-term EPF publications, whether they relate to the DAM or the BM. In doing so, we broadly classify the publications as being either statistical, ML or DL approaches, with particular attention to DL ones, given their recent prevalence. In this work, we consider ML approaches all the AI techniques that do not involve DL. A broader picture of the existing EPF literature can be found in \cite{weron2014electricity} and \cite{nowotarski2018recent}.

\subsection{Day Ahead Market Forecasting}
Literature regarding the forecasting of the DAM has received considerable attention in the past. 
\cite{lago2018forecasting} covered multiple markets, presenting an extensive range of statistical, ML, and DL models, particularly hybrid DL models. Findings indicated that \emph{deep neural networks (DNNs)}  exhibit superior performance, closely trailed by hybrid, RNN, and ML models, with statistical models lagging behind. The authors suggested that linear statistical models or their variants are likely inadequate in DAM \& BM EPF due to potential non-linearities in explanatory datasets (e.g., supply-demand curves in short-term markets with abrupt jumps \& transitions). 

RNNs have received much attention in time series forecasting due to their ability to hold/incorporate relevant information from past inputs when generating forecasts, e.g. \cite{anbazhagan2012day,chen2019brim}. LSTMs and GRUs, in particular, operate differently from standard RNNs in their ability to forget or persist certain information across time steps. For example, \cite{ugurlu2018electricity} compared DNN and statistical methods for electricity price forecasting, highlighting GRU models' superiority, particularly in Turkish DAM predictions. Lagged price values and exogenous variables improved GRUs' accuracy consistently across months. The study emphasized data availability and uncertainty estimation. Notably, three-layer GRUs outperformed other models. ANNs and LSTMs displayed high accuracy compared as well.

In \cite{lago2021forecasting}, the \emph{LASSO Estimated Auto Regressive (LEAR)} model stood out due to its low computational complexity and ease of implementation, alongside its accuracy.  Results highlighted that Ensembles of LEAR models and ensembles of DNN models produced the most precise forecasts, outperforming any of the individual models. Their research underscored the considerably higher computational cost of the DNN model compared to LEAR, despite only marginal improvement in overall forecast accuracy. The paper not only outlined best practices and reference datasets for DAM EPF but also offers a LEAR implementation, transparent codebase, and a DNN application in the DAM context.

Recently, \cite{li2021day} explored LSTM-based forecasting of Nordic DAM, emphasizing feature selection and cross-market effects on price prediction. The study examined hybrid LSTM architectures for EPF, highlighting the impact of different feature selection methods on model performance. The evaluation involved statistical measures and Diebold-Mariano tests, revealing the significance of selection methods, superiority of specific autoencoders, and the value of cross-border market features for precise predictions. The research underscored EPF's importance for power markets and policy implications for European cross-border trading enhancement.

\subsection{Balancing Market}
In a Belgium electricity market setting, \cite{dumas2019probabilistic} introduced a new probabilistic method for predicting imbalance prices in the energy market. This approach used historical data to estimate net regulation volume transition probabilities, which are then used to forecast imbalance prices based on reserve activation and marginal prices. The method outperformed other techniques like Gaussian Processes and Multi-Layer Perceptron in terms of probabilistic error measures. Further improvements and integration into bidding strategies are suggested for future work.
  
\cite{lucas2020price} employed  \emph{Gradient Boosting}, \emph{Random Forest (RF)} and \emph{Extreme Gradient Boosting (XGB)} for BM price forecasting in Great Britain, achieving accurate peak predictions but with limitations in predicting price bottoms, with XGB coming out on top. Notably, this study lacked a comparison of deep learning approaches.

Recently, \cite{narajewski2022probabilistic} focused on predicting extreme price spikes in a volatile electricity market using short-term forecasting of BM prices. Various models have been explored: Naive, LASSO, Gamlss, and Probabilistic Neural Networks. Evaluation metrics include CRPS, MAE, and empirical coverage. Findings showed Naive and Gamlss models as strong performers, with Naive having the lowest errors, albeit with poor empirical coverage. Combining Naive and Gamlss slightly enhanced coverage. They found that market volatility has a negative impact on LASSO and normal distribution-based models' effectiveness.

\section{Electricity Market Structure}\label{electricitymarketstructure}

European electricity markets differ from country to country; however, recent years have seen a significant degree of convergence in the structure and operation of short-term electricity markets with several jurisdictions following a similar template \cite{newbery2016benefits}, driven by the growing amount of variable renewable generation \cite{martinez2016impact}. Therefore, we explain the configuration of a standard European market using the example of the \emph{Irish-single electricity market} (I-SEM).

I-SEM serves as the wholesale electricity market for both Ireland and Northern Ireland. Forward energy markets span approximately 4 years to one month prior to delivery, while \emph{spot} markets enable adjustments nearer to delivery. These spot markets include:

\begin{itemize}
    \item \textit{Day-Ahead Market} (DAM): The DAM consists of one pan-European auction at noon CET for 24 hours of the next day.
    \item \emph{Intra-Day Market} (IDM): After DAM clearance, additional trading opportunities include three scheduled auctions (IDA1/IDA2/IDA3) and a continuous market.
    \item \textit{Balancing Market} (BM): The price is determined by averaging energy balancing instructions issued by the system operator for a 5-minute imbalance pricing period in I-SEM, applied over a 30-minute settlement period.
\end{itemize}

The I-SEM also includes \textit{capacity} and \textit{ancillary service} markets, similar to other jurisdictions. However, these markets are beyond the scope of this paper.

\begin{figure}
\centerline{\includegraphics[width=1\linewidth]{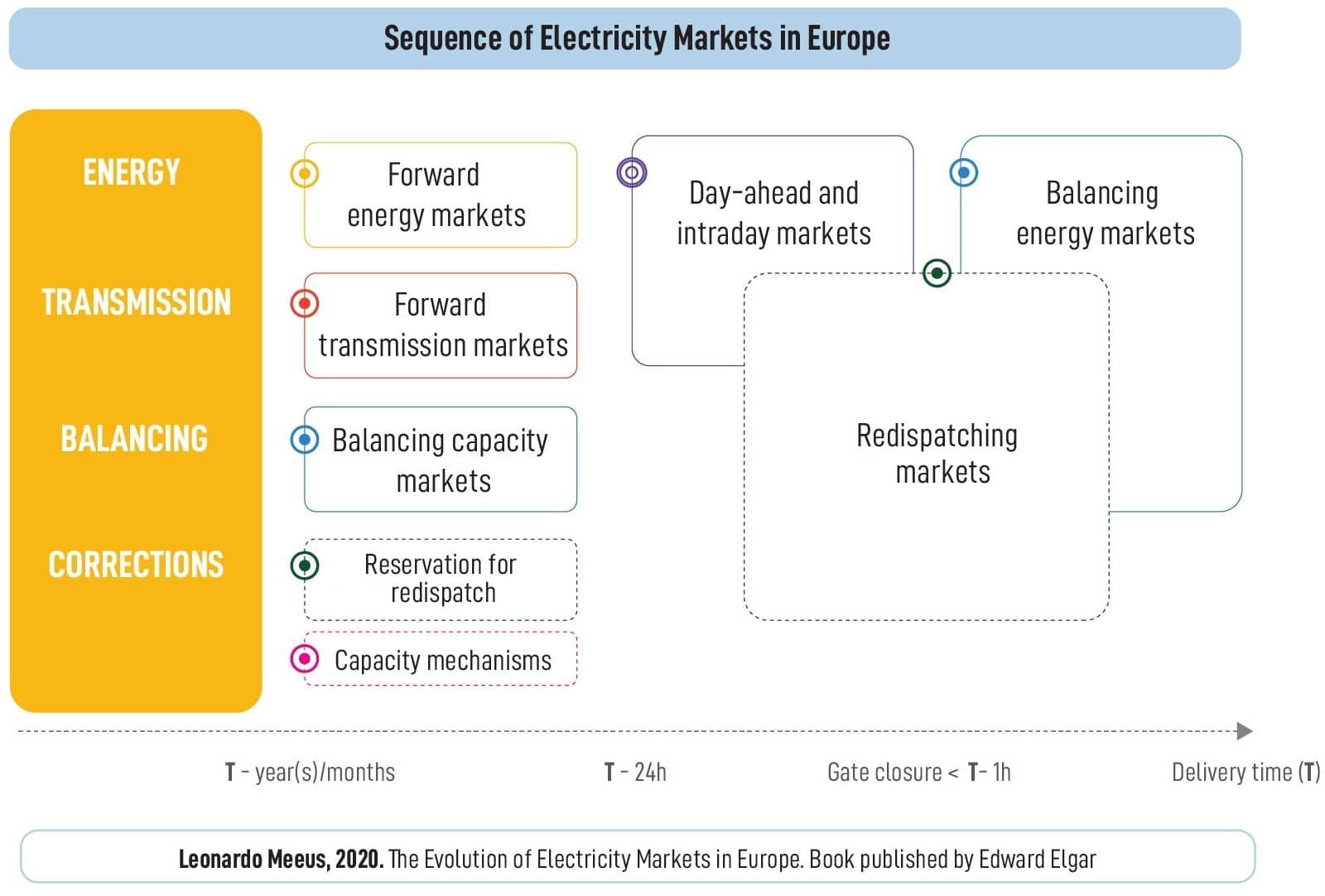}}
\caption{Schematic overview of the typical sequence of existing electricity markets in the EU. Markets in dotted lines are optional \cite{meeus2020evolution}}
\label{fig:elecfig1}
\end{figure}

The primary purpose of the BM market is to enable the system operator to perform balancing actions (i.e. matching demand and supply close to or in real-time). Hence, the BM is an important component of short-term electricity markets in a number of jurisdictions. Some features of the balancing market include:
\begin{itemize}
    \item \textit{Price Quantity Pairs}: These are the bids and offers that BM participants submit to the system operator ensuring that they are compensated for any instructed deviations to their physical notifications (e.g. their contracted positions in the ex-ante markets). 

    \item \textit{Flagging and Tagging}: The process by which the system operator excludes non-energy actions when calculating the BM price. 

    \item  \textit{De Minimis Acceptance Threshold} (DMAT): Energy actions below the DMAT threshold are excluded to not unduly impact the BM price.
    
\end{itemize}
For a more in-depth discussion of the BM, we refer the reader to \cite{bharatwaj2018real}.

\section{Analytical Framework}\label{analyticalframework}
This section presents in detail the framework used. In particular, the following sub-section presents the data sources and preprocessing, section \ref{sec:hyperparameter} describes the dataset split and hyperparametrisation procedure, and finally, in section \ref{performancemetrics}, the metrics and the statistical test used are introduced.

\subsection{Dataset}\label{datasetsanddatapreparation}
The data was sourced from the  \href{https://www.sem-o.com/}{SEMO}\footnote{https://www.sem-o.com/} and \href{https://www.semopx.com/market-data/}{SEMOpx}\footnote{https://www.semopx.com/market-data/} websites, comprising historical and forward-looking data dating from 2019 to 2022. Our goal in this study is to predict the BM price for the next 16 open settlement periods. We choose this forecast horizon explicitly to align with the availability of the input explanatory variables.

Regarding the explanatory variables, they can be categorized into two types: \textbf{Historical data} and \textbf{Forward/future-looking data}. Historical data, like BM prices, utilizes the 48 most recent BM prices to predict the next 16. In contrast, future-looking data includes forecast wind and demand (the forecast values relate to the BM settlement periods for which we are interested in predicting the price). It is possible for some explanatory variables, such as DAM prices, to appear in both a historical and future-looking context.

In our problem setting, the target is the BM price (BMP) for the proceeding, \textbf{open}, 16 settlement periods. 
\begin{align}
    Y_t=[BMP_{t+2},...,BMP_{t+17}].
\end{align}
The forecast horizon begins at $t+2$ instead of $t$, as at time $t$ when a forecast is being generated, balancing market periods $t$ and $t+1$ are closed. If a participant wants to adjust its BM commercial order data, it can only do so for BM periods $t+2$ onwards, hence our focus on the $[t+2,...,t+17]$ horizon.

The historical data considered is:
\begin{itemize}
    \item \textit{Balancing Market Prices} (BMP): This holds BM prices from the most recent \& available 24 hours, i.e. $[BMP_{t-51},...,BMP_{t-3}]$.
    \item \textit{Balancing Market Volume} (BMV): Again, the most recent 48 observations of BM volume. It is given by $[BMV_{t-51},...,BMV_{t-3}]$.
    \item \textit{Forecast Wind - Actual Wind} (WDiff): This is the difference between forecast \& actual wind data for the most recent 48 available settlement periods. $[WDiff_{t-51},...,WDiff_{t-3}]$.
    \item \textit{Interconnector Values} (I): Interconnector flows from the previous 24 hours. $[I_{t-50},...,I_{t-2}]$.
    \item \textit{DAM prices} (DAM): DAM prices from the previous 24 hours. DAM prices are published at an hourly granularity; as a result, we use the hourly price that is available for each half-hour settlement period. $[DAM_{t-48},...,DAM_t]$.
\end{itemize}

The future-looking data provided by the system is as follows:
\begin{itemize}
    \item \textit{Physical Notifications Volume} (PHPN): Sum of physical notifications for forecast horizon. That is $[PHPN_{t+2},...,PHPN_{t+17}]$.
    \item \textit{Net Interconnector Schedule} (PHI): Interconnector schedule for the forecast horizon $[PHI_{t+2},...,PHI_{t+17}]$.
    \item \textit{Renewable forecast} (PHFW): Transmission system operator renewables forecast (for non-dispatchable renewables) for the forecast horizon. \\ 
    $[PHFW_{t+2},...,PHFW_{t+17}]$.
    \item \textit{Demand forecast} (PHFD): Transmission system operator demand forecast for the forecast horizon. $[PHFD_{t+2},...,PHFD_{t+17}]$.
    \item \textit{DAM prices} (DAM): DAM prices for the next 8 hours.\\ $[DAM_{t+1},...,DAM_{t+16}]$.

\end{itemize}

For the historical data, the different time indices/intervals are solely due to data availability (i.e. the most recent and available 48 observations depend on the data source). 


\subsection{Experimental Configuration} \label{sec:hyperparameter}
In our analysis, the approach involves three main steps:
\begin{enumerate}
    \item Hyperparameter tuning
    \item Model training
    \item Model forecast (on unseen data)
\end{enumerate}
For each encountered model, we fit it using training datasets spanning 30, 60, 90, and 365 days. In model fitting and forecasting, we employ a walk-forward validation approach, which is a frequently used technique in short-term price forecasting \cite{morke2019marcos}. In our configuration, the model is retrained every 8 hours to incorporate the most recent data available.
The hyperparameters for ML or DL models are determined using the performance of the validation sets. We used $25\%$ of the training set as validation, capped at a maximum of 30 days for the 365 days dataset. 

To accommodate diverse hyperparameter configurations for validation and training sets, we repeat step 1 (hyperparameter tuning) every quarter for both DL and ML. The process is as follows:
\begin{enumerate}
    \item Given a model of interest and a starting time point $t$, obtain a validation set and determine hyperparameter settings.
    \item Using hyperparameter settings from step 1, fit the model with training data.
    \item Utilize the fitted model from step 2 to forecast using the test set.
    \item Repeat steps 2 and 3 until time point $t+90$ days is reached (i.e. walk-forward validation).
    \item Return to step 1, set $t = t+90$ days, and repeat the process.
\end{enumerate}

When performing the hyperparameter optimisation step, we use various libraries/toolboxes. For ML models such as  RF, XGB, and SVR we utilised the \textit{Scikit-learn} library \cite{pedregosa2011scikit}; for XGB type models we utilised the \textit{XGBoost} library \cite{chen2016xgboost}. For the DL models (SH DNN and MH DNN) we used \textit{Tensorflow} \cite{chollet2021deep} in conjunction with Talos \cite{yang2020hyperparameter}. For statistical model LEAR, we used the \textit{epftoolbox} \cite{lago2021forecasting}. For ML and DNN hyperparameter searches, a broad range of values was tested to find optimal settings for each time period and training size.






\subsection{Metrics}\label{performancemetrics}
Commonly encountered metrics in the EPF literature used to evaluate forecast accuracy include Mean Absolute Error (MAE) and Root Mean Squared Error (RMSE) metrics \cite{ugurlu2018performance}, as well as the symmetric mean absolute percentage error (sMAPE) metric \cite{makridakis1993accuracy}. sMAPE is preferred over MAPE due to its effectiveness in handling the frequent occurrence of prices near zero in the BM. It is defined as:

\begin{equation}\label{sMAPEEQ}
    sMAPE=\frac{100}{n} \sum_{k=1}^{n} \frac{|y_k -\hat{y}_k|}{(|y_k|+|\hat{y}_k|)}.
\end{equation}

When comparing performance metrics of models, the Diebold-Mariano (DM) test is commonly used in EPF literature to quantify statistical distinctions \cite{diebold2002comparing}. The DM test assesses null hypotheses, rejecting $H_0$, which indicates that a model has a statistically significant accuracy improvement over another.

\section{Models}\label{modelssec}
In this section, we detail the models being used to forecast BM.
We considered the best-performing models from the recent DAM EPF literature, \cite{lago2018forecasting,  lucas2020price, li2021day, lago2021forecasting}. For a more interested reader, the peculiarities of the models' implementation and hyperparameters range can be found in the paper's GitHub repository.
A high-level categorisation of the models is the following:
\begin{itemize}
    \item \textit{Naive}: This baseline is a simple approach that utilizes the preceding 8 hours of BM prices as a forecast for the subsequent 8 hours. This strategy shares similarities with the method in \cite{8472762}, which employed the DAM values from the same days a year prior as a seasonal naive DAM forecast.
    \item  \textit{LASSO Estimated AR} (LEAR): It is a modified autoregressive time series approach that integrates LASSO regularization for enhanced performance and feature selection. In linear regression, the response variable $Y_{t}$ at time point \textit{t} is formed through a linear combination of \textit{n} predictors. Variations of this model involve introducing a regularization term, such as the \textit{Least Absolute Shrinkage and Selection Operator} (LASSO) \cite{tibshirani1996regression}, or its extension, the elastic net \cite{zou2005regularization}. We adapted the implementation designed in \cite{lago2021forecasting} to the BM structure of the input/output data. Similar to \cite{lago2021forecasting}, we performed daily hyperparameter tuning for LEAR, estimating $\lambda$ with the LARS method and in-sample AIC. The optimal $\lambda$ from LARS was then used to re-calibrate LEAR through traditional coordinate descent.

    \item \textit{Random Forest} (RF): It is an ensemble model that combines several regression trees to generate predictions. It is based on the principle of bagging - short for bootstrap aggregating, splitting the data into several subsets and choosing subsets randomly with replacement, thus combining individual models with low bias and high variance, which, when aggregated, reduce the variance and retain low bias.
    \item \textit{Support Vector Regression} (SVR): It performs a nonlinear mapping of the data to a higher-dimensional space where linear functions are used to perform regression. SVR performs well in high-dimensionality space \cite{drucker1996support}. In our case, the radial basis function kernel was chosen every time.
    \item \textit{Extreme Gradient Boosting} (XGB): Similarly to the RF, XGB combines several regression trees to generate forecasts, but it is based on the principle of boosting - where learners are learned sequentially, combining models with high bias and low variance to reduce the bias while keeping a low variance.
    \item \textit{Single-Headed DNN Model} (SH-DNN): A simple extension of the traditional MLP, the ﬁrst DL model for predicting DAM prices is a DNN used in \cite{lago2018forecasting}. Its adaptation to the BM market is described in the following section.
    
    \item \textit{Multi-Headed RNN DNN Model} (MH-RNN): A non-sequential network that incorporates historical and future-looking data using dense and recurrent layers. Section \ref{MHLSTMDNNmodel} contains a more in-depth description of the model. 
\end{itemize}

\subsection{Single-Headed DNN model}\label{SHDNNmodel}
The chosen format for the single-headed DNN model is sequential, allowing for further hyperparameter optimisation with a range for the number of layers to be added for each block (where a block is comprised of between 1 to 3 connected layers) and activation function. 

The first layer flattens the data so the input data, followed by several dense layers, including a single dropout layer after block 3. Their output is fed to a series of DNN blocks, similar to that of \ref{MHLSTMDNNmodel} as seen in \ref{fig:MH_RNNDNN_fig1}. A popular choice overall for all models was the ReLU activation function, falling in line with \cite{goodfellow2016deep} findings for DL models. 

\subsection{Multi-Headed LSTM-DNN model}\label{MHLSTMDNNmodel}
The second DL model for DAM price prediction is a hybrid forecaster that merges LSTM and DNN architectures. This hybrid approach aims to incorporate a recurrent layer for capturing sequential patterns in time series data, along with a standard layer focused on the future-looking data, as in \cite{lago2018forecasting}.

\begin{figure}
\centerline{\includegraphics[width=0.6\linewidth]{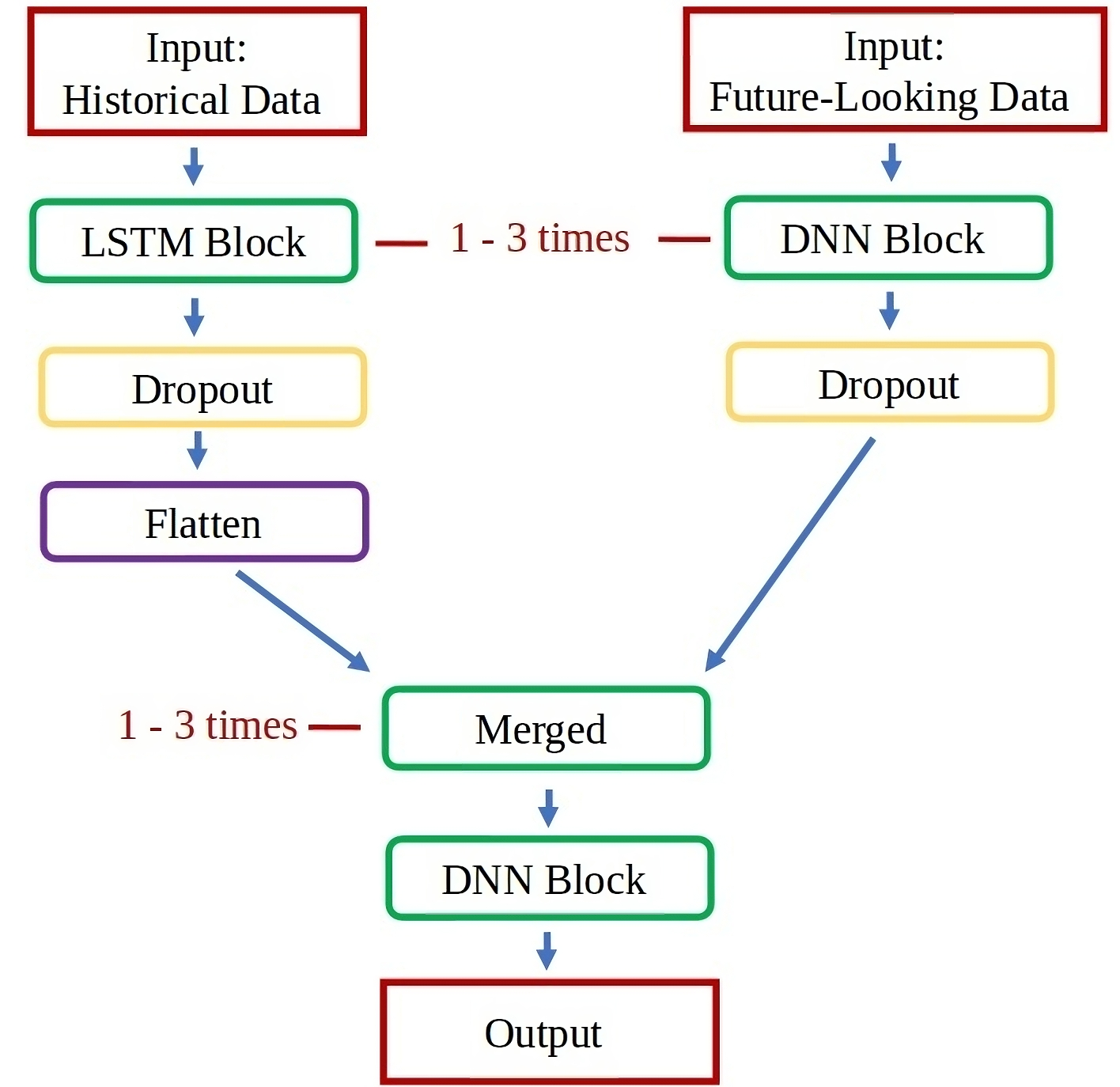}}
\caption{MH RNN DNN Model}
\label{fig:MH_RNNDNN_fig1}
\end{figure}

The LSTM branch models the historical inputs, while the DNN one takes the future-looking one. These branches have their output concatenated into one vector, and this vector is fed into a regular output dense layer containing 16 neurons for the prediction of 8 hours (or 16-time stamps). For clarity on the outline of the network, see Figure \ref{fig:MH_RNNDNN_fig1}. The obtained optimal hyperparameters for the Multi-Headed RNN DNN are 3 layers for both the LSTM and DNN networks, making for an 8-layer deep network with the concatenated and final layer. A more detailed structure of the network configuration can be found in the shared repository. 

Both DL Models include dropout layers and early stopping, known to be effective in larger deep models for countering overfitting \cite{chollet2021deep}. We trained the models for 300 epochs; however, if there is no improvement in the validation error for 30 epochs, the training halts.

\section{Results}\label{resultssec}
In this section, we present an in-depth analysis of the performance of the models presented above. The first part focuses on the accuracy of the prediction, also comparing the impact of different training horizons. The results are examined in relation to the performances on the DAM. The models are then compared to see if their differences are statistically significant. Later, we analyse how the prices and their predictions change during the day to understand which parts of the BM are harder to forecast. Finally, we analyse the models in terms of the computational effort required.

\subsection{Accuracy}
Table \ref{table:MAE} offers a comprehensive overview of MAE, RMSE, and sMAPE errors for different models and training sizes. 
Comparisons are also made with DAM metrics for SH DNN and LEAR models over identical time frames, with noteworthy differences emerging between DAM and BM predictions.

\setlength{\tabcolsep}{2.1pt}
\begin{sidewaystable*}\centering

\begin{tabular}{@{}rrrrrrcrrrrrcrrrrr@{}}\toprule
& \multicolumn{5}{c}{MAE} & \phantom{a}& \multicolumn{5}{c}{RMSE} &
\phantom{a} & \multicolumn{5}{c}{sMAPE}\\
\cmidrule{2-6} \cmidrule{8-12} \cmidrule{14-18}\\
\textbf{Training Size} 
& 30 days & 60 days & 90 days  & Max Data & DAM && 30 days & 60 days & 90 days  & Max Data & DAM && 30 days & 60 days & 90 days  & Max Data & DAM\\ \midrule
\textbf{Model} \\
Naive & 51.96 & 51.96 & 51.96 & 51.96 & && 84.27  & 84.27 & 84.27 & 84.27 &  &&  &  &  &  & \\
LEAR & 35.95 & \textbf{34.12} & \textbf{33.57} & \textbf{32.82} & 10.15 &&  \textbf{57.89} & \textbf{56.38} & \textbf{56.34} & 56.53 & 18.84 && 72.14 & \textbf{67.19} & \textbf{66.20} & \textbf{64.93}& 22.68\\
XGB & \textbf{35.35} & 34.40 & 34.29 & 33.71 & &&  60.46 & 58.03 & 56.89 & 57.03 & && 69.71 & 67.62 & 67.12 & 66.89& \\
RF & 36.15 & 35.24 & 34.92 & 34.66 & &&  58.62 & 57.27 & 56.66 & \textbf{56.22} & && \textbf{68.67} & 68.55 & 68.45 & 67.18& \\
SVR & 41.43 & 38.01 & 37.70 &  37.39 & &&  69.81 & 66.12 & 65.58 & 64.92 & && 78.35 & 69.85 & 68.97 & 68.09 & \\
SH DNN & 40.14 & 38.67 & 38.31 & 37.41 & \textbf{7.34} &&  64.13 & 68.91 & 62.25 & 61.95 & \textbf{13.86} && 77.85 & 72.10 & 72.05 & 70.81 & \textbf{19.01} \\
MH RNN DNN & 41.42 & 40.89 & 40.35 & 38.28 & &&  64.83 & 64.02 & 64.42 & 61.96 & && 79.77 & 77.20 & 76.68 & 74.84 & \\
\bottomrule
\end{tabular}
\caption{Balancing Market prediction accuracy for the different training sizes. In bold, the best-performing model. MAE is measured in  \euro/MWH.}
\label{table:MAE}
\end{sidewaystable*}

\begin{figure}
    \centering
        \includegraphics[scale=0.27]{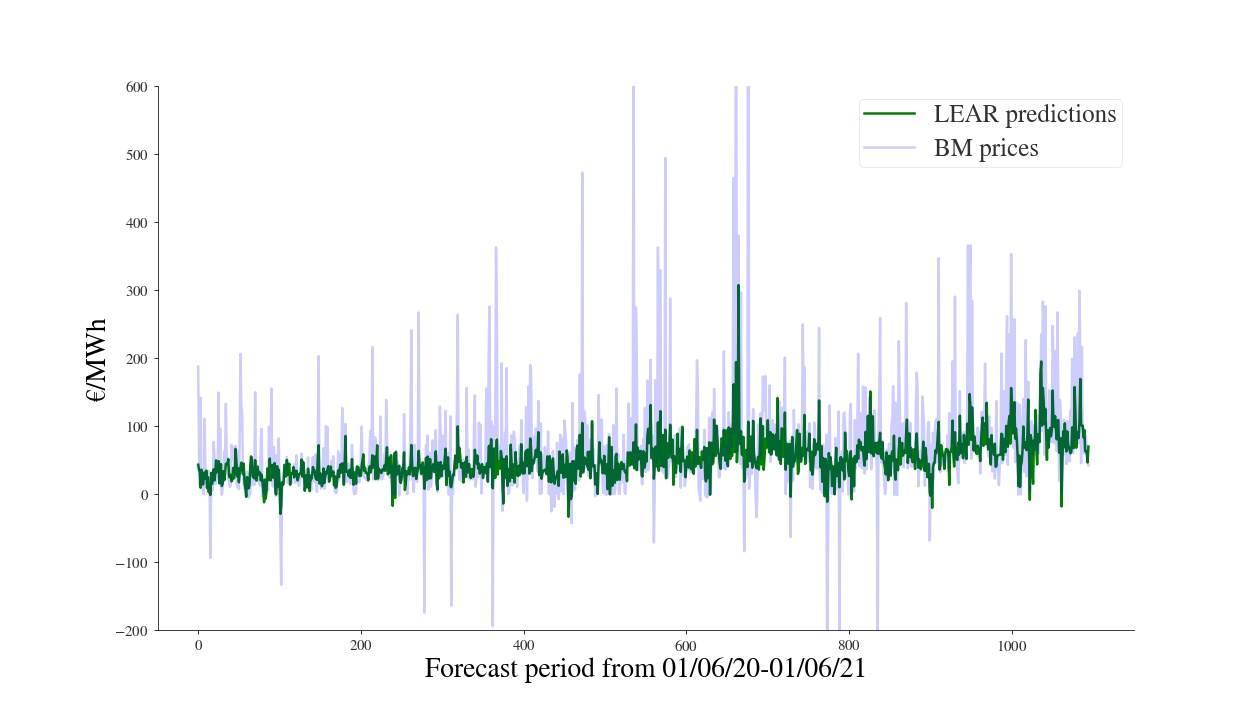}
        \caption{LEAR Forecast for BM}
        \label{fig:LEAR}
\end{figure}

LEAR emerges as the top-performing model for the BM, with limited exceptions for shorter training periods. Its subtle adjustments led to relatively smoother and lower variation in the predictions. Figure \ref{fig:LEAR} displays a comparison between the BM prices and LEAR's predictions. We can see that the model does not frequently attempt to predict significant spikes. This seems to be the reason for its good accuracy; the model returns forecasts that might not consistently predict the exact direction of BM prices but avoid producing highly inaccurate predictions. Notably, increasing the training set size had a significant impact on LEAR's performance. The MAE decreased by \euro3.31/MWh when transitioning from a 30-day training period to 310 days. 

ML models, including RF, XGB, and SVR displayed varied performance. XGB stood out by closely approaching the accuracy of LEAR and outperforming RF and SVR. Both XGB and RF outperformed the DL models, while SVR struggled to capture data patterns. This finding aligns with the UK BM paper \cite{lucas2020price}, confirming XGB's superiority over gradient boosting and RF models in a similar setting. 

\begin{figure}[htbp]
    \centering
    \includegraphics[scale=0.27]{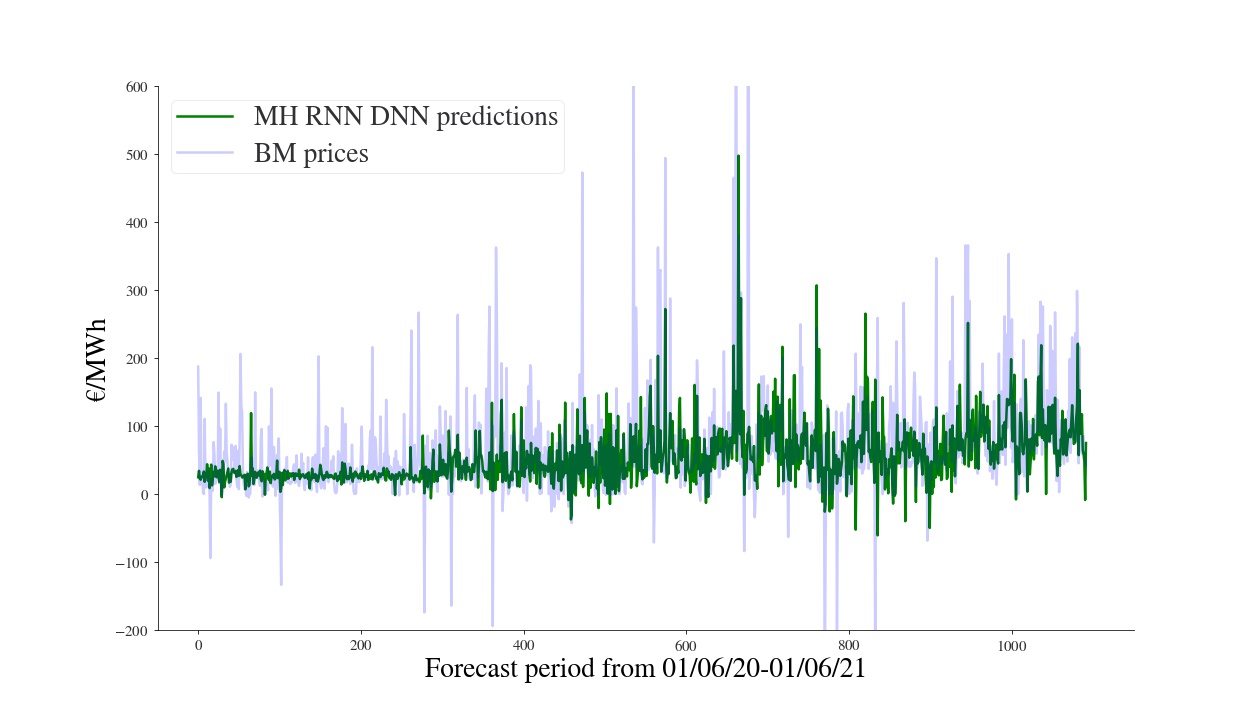}
    \caption{MH RNN DNN Forecasts for BM}
    \label{fig:MH_DNN}
\end{figure}

DL models exhibit inferior performance, consistently underperforming across all metrics, unsurprising considering their struggles in \cite{dumas2019probabilistic,narajewski2022probabilistic}. While these models perform admirably in domains like the DAM, their effectiveness in predicting BM encounters challenges, including overfitting to recent price spikes and difficulties with complex BM dynamics. This divergence underscores the complexities inherent in electricity market forecasting. Figure \ref{fig:MH_DNN} shows the performances on part of the test set. Comparing them to LEAR's (Figure \ref{fig:LEAR}), we can see that the DL model behaves overconfidently, predicting considerable spikes. If these predictions are incorrect, they are strongly penalised by the metrics compared to more cautious ones. The SH DNN model outperformed the MH RNN DNN model, showing improvements as the training set size increased (resulting in a \euro2.47/MWh reduction in MAE). While the SH DNN model displayed better forecasting ability than XGB and LEAR during the least volatile period (first 3 months), its performance deteriorated as market volatility increased.
The MH RNN DNN model, which exhibited excellent results in the DAM \cite{lago2018forecasting}, struggled to accurately forecast BM prices. This could be attributed to the current dataset's unsuitability for the split approach used, where data is fed into two separate branches. If past and future data heavily influence each other, both structures may fail to establish these relationships. In contrast, the SH DNN model avoids assumptions about input data, enabling more comprehensive relationship-building. 

We investigate the impact of training data size on model performance. Having access to a higher amount of data strongly benefitted all the models. Increasing the training data window to 365 days resulted in an average decrease of 3.00 MAE, 2.67 RMSE, and 5.33 sMAPE across models. Surprisingly, the DL models had a smaller benefit from the increased amount of data. The consistent enhancement of model performance highlights the importance of historical data.

\subsubsection*{Statistical Test}
The differences in accuracy can be caused by the intrinsic stochasticity of the dataset. For this reason, a test comparing them in a statistically significant manner provides more robust insights. Table \ref{table:DB} presents the results of the Diebold-Mariano (DM) test on the MAE, where column (model 2) and row (model 1) headers are organized by decreasing MAE errors. The \cmark symbol signifies a statistical significance of 95\% or more in accepting the alternative hypothesis from the DM test; otherwise, the comparison is marked with \xmark. This implies that model 1's predictive accuracy is statistically significantly superior to that of model 2. 

The models' performances can be grouped into three groups. LEAR and the two ensemble ML models (XGB and RF) have similar results, significantly outperforming the others. The SVR and DL networks differences are not consistent enough to accept the alternative hypothesis with reasonable confidence. Finally, all the models considered are significantly better than a naive baseline.

\begin{table*}[ht]
\centering
\renewcommand{\arraystretch}{0.8}
\setlength{\tabcolsep}{6pt} 
\small
\begin{tabular}{@{}r*{7}{c}@{}}\toprule
& \multicolumn{7}{c}{DB Test - MAE} \\
\cmidrule{2-8} 
\textbf{Model} 
& LEAR  & XGB & RF & SVR & SH DNN & \multicolumn{1}{c}{MH RNN DNN} & Naive \\ 
\midrule
 
LEAR  &  &  \xmark &  \xmark &  \cmark & \cmark  & \cmark & \cmark \\
 
XGB   &   &  &  \xmark &  \cmark & \cmark  & \cmark & \cmark \\
 
RF    &   &  &   &  \cmark & \cmark  & \cmark & \cmark \\
 
SVR  &  &  &  &   & \xmark & \xmark &  \cmark\\
 
SH DNN   &  &  &  &   &  &\xmark  & \cmark \\
 
MH RNN DNN  &  &  &  &   &  &  &  \cmark \\
 
Naive &  &  &  &  & &   &   \\

\bottomrule
\end{tabular}
\caption{DB test statistical significance}
\label{table:DB}
\end{table*}

\subsubsection*{Hourly Analysis}
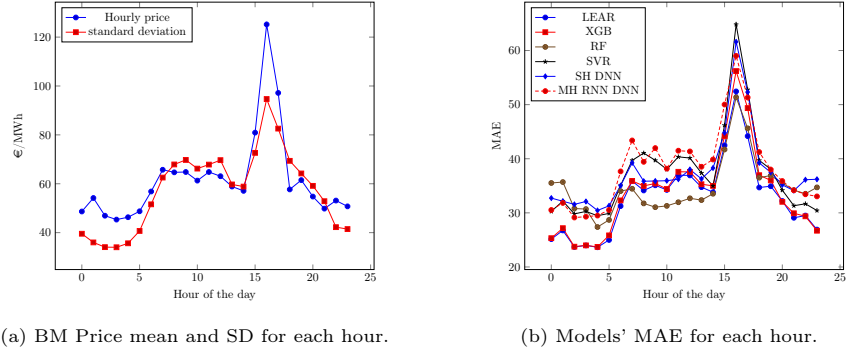
\begin{figure}
    \centering
    \begin{subfigure}{0.48\textwidth}
        \centering
        \begin{tikzpicture}[scale=0.45]
            \begin{axis}[
                xlabel={Hour of the day},
                ylabel={\euro /MWh},
                legend style={at={(0.02,0.98)}, anchor=north west}
            ]

        \addplot coordinates{
        (0, 48.65005712422422) (1, 54.16913994724424)    (2, 46.92563410510371)    (3, 45.34076000616414)    (4, 46.30820821572427)    (5, 48.734717729785956)    (6, 56.83225199669099)    (7, 65.73110973312967)    
        (8, 64.67472650472565) (9, 64.80942237125191)    (10, 61.287690990707226)    (11, 64.80889713147425)    (12, 63.08194237819602)    (13, 58.868838332757505)    (14, 57.09658070090733)    (15, 80.9201850116529)  
        (16, 125.18935245918453) (17, 97.17178529433059)    (18, 57.681863136041294)    (19, 61.49276649352058)    (20, 54.7584139043862)    (21, 49.82232106925096)    (22, 53.12247904018533)    (23, 50.74760158155792) 
        };

        \addplot coordinates{
        (0, 39.54687671232877) (1, 35.99624657534246)    (2, 34.06289041095891)    (3, 34.01169863013699)    (4, 35.6353698630137)    (5, 40.69364383561644)    (6, 51.579109589041096)    (7, 62.532931506849316)    
        (8, 67.89793150684932) (9, 69.74671232876713)    (10, 66.20098630136987)    (11, 67.8407397260274)    (12, 69.65676712328768)    (13, 59.72190410958905)    (14, 58.839410958904104)    (15, 72.64169863013699)  
        (16, 94.66397260273972) (17, 82.56198630136986)    (18, 69.38217808219177)    (19, 64.21502739726027)    (20, 59.143068493150686)    (21, 52.8613287671233)    (22, 42.24069863013698)    (23, 41.455602739726025) 
        };
          \legend{Hourly price, standard deviation}
         \end{axis}
        \end{tikzpicture}  
        
        \caption{BM Price mean and SD for each hour.}
        \label{fig:HourlyPrice&SD}
    \end{subfigure}
    \hfill
    \begin{subfigure}{0.48\textwidth}
        \centering
        \begin{tikzpicture}[scale=0.45]
            \begin{axis}[
                xlabel={Hour of the day},
                ylabel={MAE},
                legend style={at={(0.02,0.98)}, anchor=north west}
            ]
        \addplot coordinates{
        (0, 25.15) (1, 26.75) (2, 23.7) (3, 23.98) (4, 23.64) (5, 24.99) (6, 31.25) (7, 35.88) (8, 34.14) (9, 35.09) (10, 34.24) (11, 36.96) (12, 36.92) (13, 34.75) (14, 33.82) (15, 42.47) (16, 52.44) (17, 44.18) (18, 34.69) (19, 34.92) (20, 32.19) (21, 29.09) (22, 29.5) (23, 26.91) 
        };
        \addplot coordinates{
        (0, 25.32) (1, 27.17) (2, 23.75) (3, 23.99) (4, 23.71) (5, 25.84) (6, 32.3) (7, 35.9) (8, 34.98) (9, 35.43) (10, 34.41) (11, 37.59) (12, 37.57) (13, 35.29) (14, 34.99) (15, 44.16) (16, 56.2) (17, 49.36) (18, 36.96) (19, 36.0) (20, 32.04) (21, 29.92) (22, 29.36) (23, 26.7) 
        };
        \addplot coordinates{
        (0, 35.52) (1, 35.69) (2, 30.8) (3, 30.7) (4, 27.38) (5, 28.69) (6, 34.02) (7, 34.45) (8, 31.75) (9, 31.05) (10, 31.29) (11, 31.96) (12, 32.69) (13, 32.36) (14, 33.53) (15, 41.71) (16, 51.36) (17, 45.62) (18, 36.5) (19, 36.9) (20, 35.5) (21, 34.2) (22, 33.53) (23, 34.71) 
        };
        \addplot coordinates{
        (0, 30.27) (1, 32.18) (2, 29.81) (3, 30.26) (4, 29.47) (5, 29.9) (6, 35.07) (7, 39.69) (8, 41.05) (9, 39.71) (10, 38.04) (11, 40.35) (12, 40.13) (13, 37.36) (14, 34.97) (15, 46.19) (16, 64.82) (17, 52.77) (18, 39.7) (19, 38.06) (20, 34.19) (21, 31.32) (22, 31.67) (23, 30.43) 
        };
        \addplot coordinates{
        (0, 32.73) (1, 32.14) (2, 31.57) (3, 32.1) (4, 30.44) (5, 31.3) (6, 35.03) (7, 39.24) (8, 35.85) (9, 35.88) (10, 35.94) (11, 36.22) (12, 38.01) (13, 36.28) (14, 38.29) (15, 44.79) (16, 61.6) (17, 52.31) (18, 39.23) (19, 37.48) (20, 35.17) (21, 34.11) (22, 36.12) (23, 36.21) 
        };
        \addplot coordinates{
        (0, 30.53) (1, 31.83) (2, 29.14) (3, 29.29) (4, 29.5) (5, 30.54) (6, 37.66) (7, 43.38) (8, 39.43) (9, 41.97) (10, 38.17) (11, 41.49) (12, 41.35) (13, 38.52) (14, 39.86) (15, 50.02) (16, 59.01) (17, 51.32) (18, 41.24) (19, 38.0) (20, 35.88) (21, 34.17) (22, 33.43) (23, 33.04) 
        };

          \legend{LEAR, XGB, RF, SVR, SH DNN, MH RNN DNN}
          \end{axis}
        \end{tikzpicture}
        \caption{Models' MAE for each hour.}
        \label{fig:HourlyMAEModel}
    \end{subfigure}
    \caption{Hourly breakdown of the BM prices and the models' forecast error.}
    \label{fig:HourlyAnalysis}

\end{figure}

Analysing the average trends over the 24 hours gives better insights into the reasons for the performance differences. Figure \ref{fig:HourlyPrice&SD} shows the average price and its standard deviation, while Figure \ref{fig:HourlyMAEModel} shows the models' MAE both aggregated by the hour. We considered the models trained with 365 days of data.  We can see that there are considerable variations in prices over the day; unsurprisingly, there is a strong peak in the 15:00 - 18:00 range. The same period is characterized by the highest variability as well. During the night, the prices and their standard deviation are considerably lower. The forecast error of the predictors follows a similar pattern, where higher prices and variabilities lead to less accurate predictions in absolute terms.
These peculiarities also strongly affect the individual model performances. The DL models are more accurate on the night prices, with less uncertainty. However, their attempt to predict the peaks leads to higher errors during the more volatile parts. On the contrary, LEAR and the best-performing ML models have a more stable error, being the most accurate over the volatile part. 
Techniques that more accurately predict lower and stable prices might be preferred over the best model on average for specific applications. For example, when having to choose which part of the night is better to charge a battery.

\subsubsection*{Comparison with DAM}
\begin{figure}
    \centering
        \includegraphics[scale=0.31]{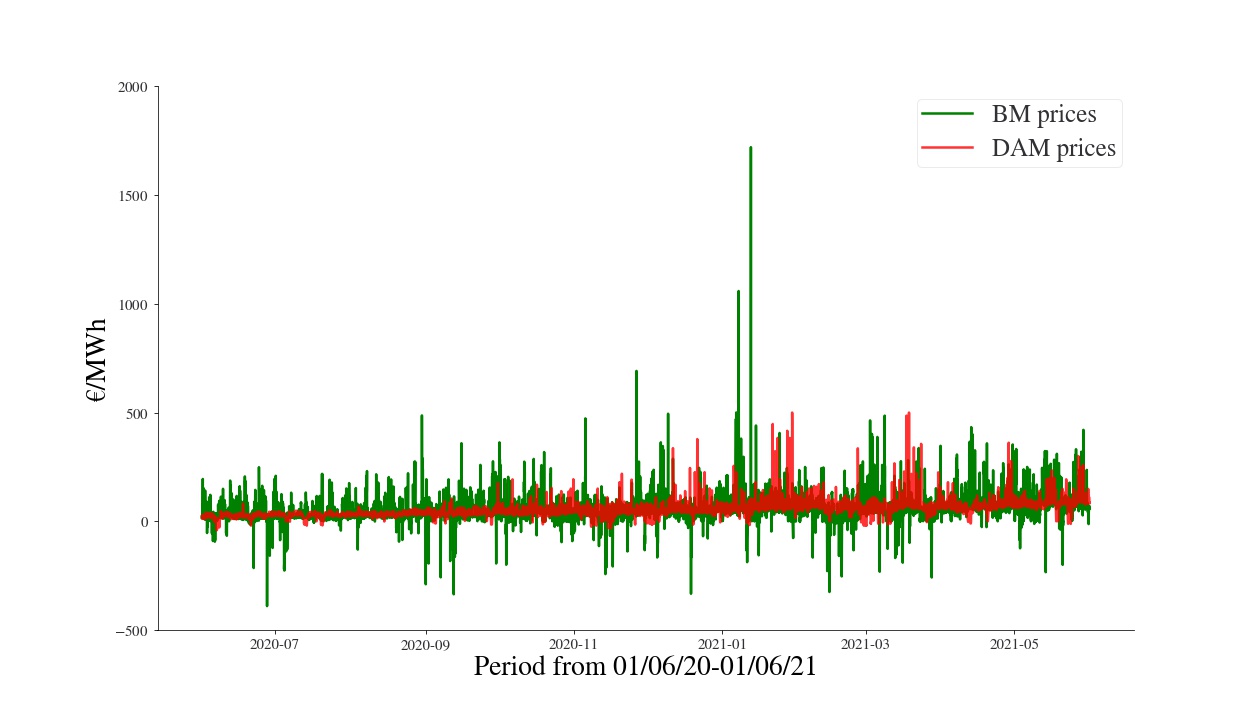}
        \captionsetup{justification=centering}
        \caption{DAM and BM prices.}
        \label{fig:DAMBMPRICES}
 
\end{figure}

The differences between the two markets are consistent. Figure \ref{fig:DAMBMPRICES} gives an idea of the difference in price ranges and uncertainty of the two markets.  Considering the substantial volatility observed in the BM, where contiguous trading periods often witness extreme price changes, both negative and positive, for example:

\begin{itemize}

  \item For the 4 BM settlement periods on 27/06/2020 in the 18:30-20:30 time horizon, prices veered from \euro2.01/MWH to \euro-390.07/MWH to \euro-38.69/MWH before ending at \euro3.12/MWH.

  \item On the 07/01/2021 15:00-16:30 prices jumped from \euro359.89/MWh to \euro1222.89/MWh, then to \euro1059.07/MWh and back to \euro320.88/MWh. For context, within the related 12-hour time block, the BM price ranged from a low of -\euro98.42/MWh to the aforementioned high of \euro1222.89/MWh.

\end{itemize}

\begin{figure}
    \centering
        \centering
        \includegraphics[scale=0.27]{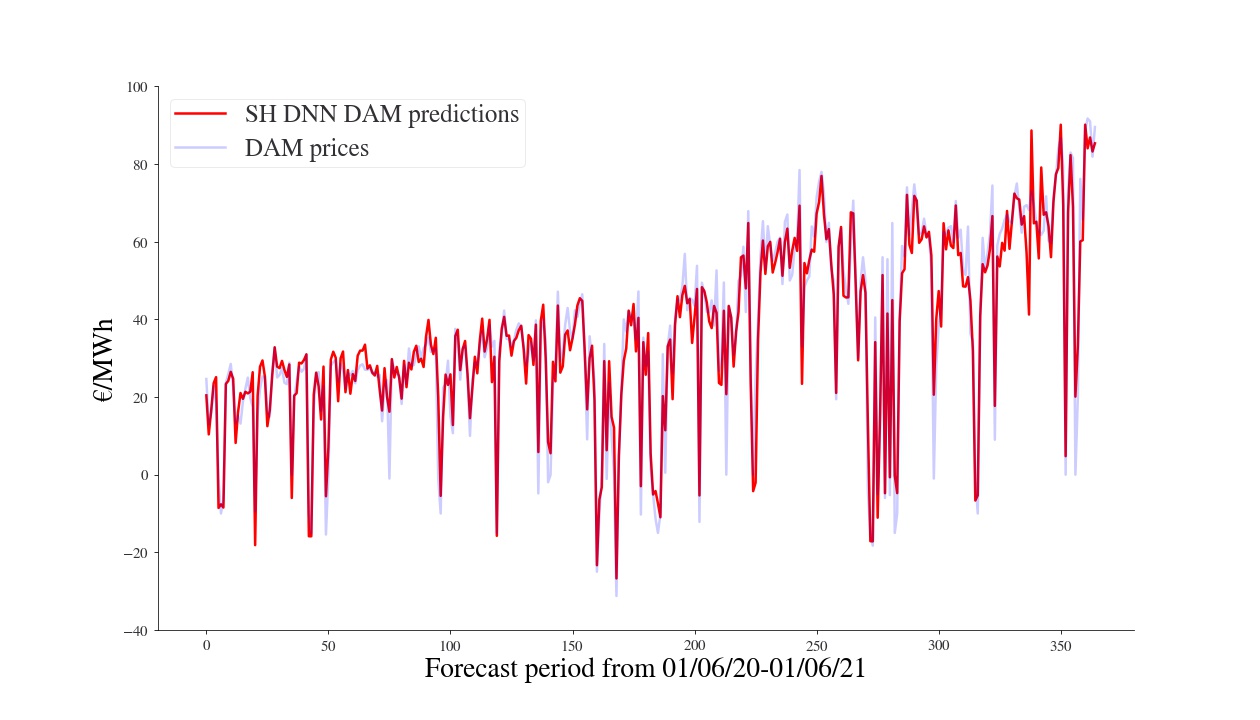}
        \caption{SH DNN Forecast for DAM}
        \label{fig:damshdnn}
\end{figure}

While the DL models encounter challenges in BM predictions, they exhibit more accurate forecasts in the DAM domain, reflecting their ability to capture price spikes and market dynamics. DL models' superior performance in the DAM compared to the BM underscores the influence of market structure and data characteristics on model performance. The SH DNN outperforms LEAR by a significant \euro2.81/MWh, confirming the results present in the literature. In Figure \ref{fig:damshdnn}, we can see the predictions of the SH DNN in comparison to the DAM prices. Compared to Figure \ref{fig:MH_DNN}, we can see that the predicted peaks are more accurate. As outlined above, the DNNs overfit to recent spikes in the BM and follow sharp increases in the BM with an increase in the next 8 hours of predictions, which is frequently wrong. The more stable DAM market is better suited for the DL approaches.  This comparison strengthens the importance of an investigation for better forecasting models tailored to the BM.

\subsection{Computational Efficiency}
In practical terms, when generating forecasts in the live setting of the BM, participants should consider the time lag between available information (explanatory variables) and the computational time required for model output. This can vary based on factors such as the deployed model and available hardware. In our analysis, DL models were trained on a GPU (Nvidia GTX 1080 ti), while statistical and ML models were trained on the CPU (Intel(R) Core(TM) i7-6800K CPU @ 3.40GHz). 
\begin{figure}[ht!]
    \centering
    \begin{tikzpicture}
        \begin{axis}[
            ybar,
            bar width=0.7cm,
            ylabel={Time (min)},
            symbolic x coords={LEAR, RF, XGB, SVR, SH DNN, MH DNN},
            xtick=data,
            enlarge x limits=0.10,
            title=Computational Cost of Each Model for 365 Days of Training Data,
            xlabel={Model},
            legend style={at={(0.5,-0.15)},anchor=north,legend columns=-1},
        ]
        \addplot+[error bars/.cd, y dir=both, y explicit] coordinates {
            (LEAR, 0.5) +- (0.1, 0.1)
            (RF, 3) +- (0.5, 0.5)
            (XGB, 0.5) +- (0.1, 0.1)
            (SVR, 10) +- (0.5, 0.5)
            (SH DNN, 3) +- (2, 2)
            (MH DNN, 18) +- (13, 12)
        };
        \end{axis}
    \end{tikzpicture}
    \caption{Computational Time Required for Training}
    \label{fig:comptime}
\end{figure}
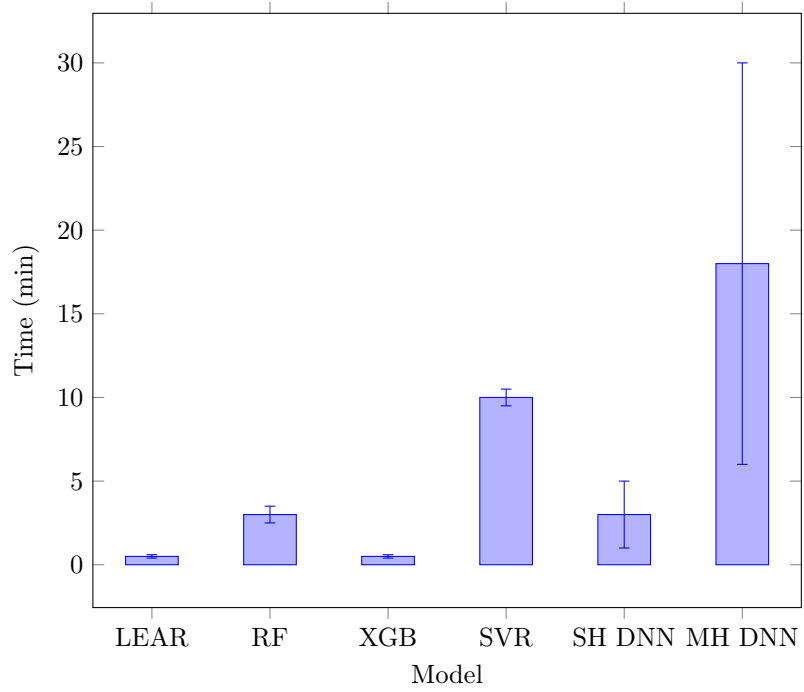

Figure \ref{fig:comptime} shows the training time required by all the models. DL models generally entail higher computational costs compared to ML and statistical counterparts. Notably, LEAR and XGB perform excellently in computational cost and forecast accuracy, making them well-suited for real-time applications.
As expected, DL model training requires a considerable amount of time and exhibits the highest variability. Many factors affect their performance, such as the number of parameters, training data size, and number of epochs. The MH model's higher complexity and higher number of parameters considerably increase the time required for its training. The plot refers to training times; the models need to be retrained quite often to incorporate the most recent training data. The prediction time for a single point of the test set is negligible for all the models.

\section{Conclusion}\label{conclusionsec}
This paper introduced a framework to benchmark predictive models for forecasting volatile electricity prices in the BM. In an extensive numerical analysis, we evaluated statistical, ML, and DL approaches that have been proven successful in the literature on similar tasks. The DL models that excelled in the more stable DAM struggled to forecast the uncertainty of the most volatile part of the BM, where simpler approaches shone. In particular, LEAR and tree ensemble techniques achieved the best accuracy due to their conservative predictions. The DL models encountered challenges in handling sudden price spikes, resulting in elevated errors. The analysis also showed interesting insights on the Irish BM market. The original dataset, the models, and the framework are provided under an open-source license. 

This BM price forecasting work suggests potential future directions. The hourly analysis suggests that an algorithm portfolio using different approaches for different parts of the day could leverage the differences between the models. Exploring additional BM markets could generalize findings and highlight differences. Some BM operators offer more granular 5-minute data, allowing an alternative modelling approach. Moreover, higher data availability would allow the deployment of transfer learning approaches.  
Further BM research avenues could encompass generating probabilistic forecasts with modern approaches (e.g. conformal prediction, \cite{kath2021conformal}), treating price prediction as a classification challenge by employing DNNs' spike detection and leveraging forecasts for improved trading opportunities, potentially in conjunction with the DAM, to enhance financial outcomes for battery energy storage systems participants.

\section*{Acknowledgments}
This work was conducted with the financial support of Science Foundation Ireland under Grant Nos. 18/CRT/6223, 16/RC/3918 and 12/RC/2289-P2 which are co-funded under the European Regional Development Fund. For the purpose of Open Access, the author has applied a CC BY public copyright licence to any Author Accepted Manuscript version arising from this submission.
\bibliography{elsarticle-template}

\end{document}